# MICROSTRIP COUPLER DESIGN USING BAT ALGORITHM

EzgiDeniz Ulker[1] and Sadik Ulker[2]

[1]Department of Computer Engineering, Girne American University, Mersin 10, Turkey
[2] Department of Electrical and Electronics Engineering, Girne American University, Mersin 10, Turkey

*ABSTRACT*

*Evolutionary and swarm algorithms have found many applications in design problems since todays computing power enables these algorithms to find solutions to complicated design problems very fast. Newly proposed hybridalgorithm, bat algorithm, has been applied for the design of microwave microstrip couplers for the first time. Simulation results indicate that the bat algorithm is a very fast algorithm and it produces very reliable results.*

## 1. INTRODUCTION

With the use of computers and high computing power nowadays, evolutionary and swarm algorithms have found applications in optimization problems and design problems. Some of these algorithms are genetic algorithms proposed by Holland [1], particle swarm optimization algorithm developed by Kennedy and Eberhart [2], immune algorithm proposed by Farmer et.al. [3] and Harmony Search Algorithm proposed by Geem, Kim and Logonathan [4]. These algorithms all can be considered artificial intelligence concepts and most of them are derived from nature and biological systems. Applications of these methods have found a great interest in engineering design such as in microwave circuit design [5]. Multi-objective optimization problems is of more interest and these methods are applied, as an example in parameter estimation of induction motors [6], in the design of low noise amplifiers [7] and in broadband microwave absorbers [8].A new hybrid method, bat algorithm, was proposed by Yang as a powerful tool which combines variety of these methods namely particle swarm optimization and harmony search algorithms [9]. In this paper, the aim is to test this algorithm on a microwave design problem. In order to achieve this aim, design of a microwave microstrip coupler is considered. For the performance, the items of accuracy and speed are considered as means of performance metrics. In the next section, the main steps of the bat algorithm are explained. In the problem definition and results sections, the microstrip coupler design problem is described and the obtained results are presented. Lastly, the conclusions section gives the concluding remarks of the paper.

## 2. ALGORITHM

Bat Algorithm is derived through the echolocation behaviour of bats. Bats are very good at finding the prey and they can distinguish very small prey and obstacles even in the dark. They can also detect the distance of the prey, type of the prey and the moving speed of the prey using the echolocation behaviour [9]. The algorithm mimics the same process of how bats can find and detect the food source with the help of echolocation. It starts by defining a purpose function *f(x)* which needs to be optimized. We create some possible candidates which form the bat population that will be used in the objective function to calculate their fitness values. Bats move randomly for finding the food source. In bat algorithm, variables in the search space such as velocity $v_i$, position $x_i$, frequency *f*, rate *r*, and loudness $A_0$ should be updated in the iteration process. As a result of updating certain variables the new bat population is created. Fitness value of each bat in the new population is calculated then sorted according to the fitness values. After certain evaluations, the fitness value that is closest to the optimal value of the objective function is the optimal solution of our problem. We can list the steps of the algorithm as follows.

Step 1: Generate a set of bat population, velocity $v_i$, position $x_i$, frequency *f*, rate *r*, and loudness $A_0$.

Step 2: Calculate the fitness value of each bat in the population.

Step 3: Sort the bat population according to fitness values of candidates.

Step 4: Update variables for bats flying randomly. Therefore new bat population is created with using the updated variables $v_i$, $x_i$, *f*, *r* and $A_0$.

Step 5: Calculate the fitness value of each bat in the new bat population, and rank them.

Step 6: Create and replace the predefined number of new random bats into the population.

Step 7: Repeat Steps 4-6 while the stopping criterion is not met.

Bat algorithm consists of similar steps with the other algorithms such as initialization, generation of new candidates in population by flying randomly, local and global search. New solutions are generated by moving virtual bats according to following expressions

$$f_i = f_{min} + (f_{max} - f_{min})\beta , \quad (1)$$
$$V_i^{t+1} = V_i^t + (X_i^t - best)f_i , \quad (2)$$
$$X_i^{t+1} = X_i^t + V^{t+1}, \quad (3)$$

where *β* is a random number in between (0, 1) and *best* is the best solution in population. $f_{min}$ and $f_{max}$ show the frequency limits and adjusted to 0 and 100, respectively. Bat algorithm uses exploitation to focus the search around the current best candidate. This is called random walk in algorithm. It modifies the current best solution and done by the following expression

$$X_{new} = X_{old} + \varepsilon A^t , \quad (4)$$

where ε is a random number in between (0, 1) and $A^t$ is the average loudness of all bats at this time *t*. The loudness $A_i$ and the rate $r_i$ of pulse emission are updated in the iteration process according to following expressions

$$A_i^{t+1} = \alpha A_i^t \,, \tag{5}$$
$$r_i^{t+1} = r_i^0 [1 - \exp(-\gamma t)] \,, \tag{6}$$

where α and γ are constants in between (0, 1) and γ >0, respectively.

## 3. PROBLEM DEFINITION

Microstrip couplers are part of many microwave circuits and components and their design is of interest. Researchers in the past used the geometry of the lines to develop a model for the even and odd mode impedances of the line and hence calculate the coupling coefficient of the microstrip line accordingly [10-12]. The theoretical foundations and the models predict very close proximity to the real experimental results that are obtained, however the complexity of equations make the design a difficult task unless an optimizing tool in a CAD program is used. In this case we will use these equations in an algorithm and then the obtained results will be used in a CAD program to observe whether the algorithm has found correct dimension values for the desired coupling. We can consider the drawing in Figure 1 as a simple microwave microstrip coupler with the unknowns indicated [13].

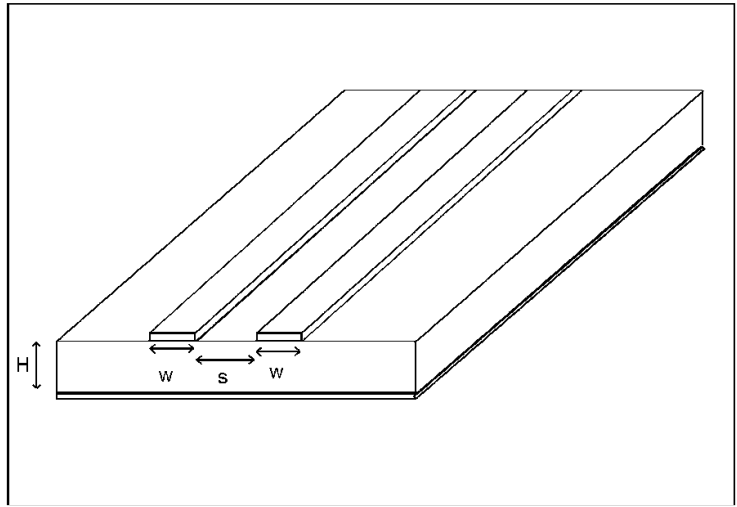

Figure 1. Microstrip Coupler [13]

In this figure, two microstrip lines of width *w* are separated by spacing *s* on a grounded dielectric with thickness *H*. Electromagnetic coupling occurs between two microstrip lines and the proper values *w*, *s* and *H* need to be optimized for desired coupling. In order to test the algorithm, the equations are established in a manner to obtain the odd and even mode impedances ($Z_{oo}$ and $Z_{oe}$) using the ratios '*whse*' and '*whso*' for the microstrip line using the equations proposed by Akhtarzad[12] as given below

$$whse = \frac{2}{\pi} \cosh^{-1}\left(\frac{2h-g+1}{g+1}\right), \tag{7}$$

$$whso = \frac{2}{\pi} \cosh^{-1}\left(\frac{2h-g-1}{g-1}\right) + \frac{4}{\pi(1+\varepsilon_r/2)} \cosh^{-1}\left(1 + 2\frac{w/H}{s/H}\right), \tag{8}$$

$$g = \cosh\left[\frac{1}{2}\pi(s/H)\right], \tag{9}$$

$$h = \cosh\left[\pi(w/H) + \frac{1}{2}\pi(s/H)\right], \tag{10}$$

where equation 8 is valid for dielectric constant ($\varepsilon_r$) materials with values less than 6. Ratios *whse* and *whso* can be used to determine $Z_{oo}$ and $Z_{oe}$ as explained by Akhtarzad et.al. [12] and hence the coupling coefficient, *C*, which can be simply obtained by using the following expression

$$C = \frac{Z_{oe} - Z_{oo}}{Z_{oe} + Z_{oo}}. \tag{11}$$

## 4. RESULTS

For test purposes we set the desired coupling coefficient to be 0.2. The substrate with dielectric constant being 3.9 was considered for the microstrip coupler. In a random manner we created 20 'bats' which held the values of *W* (*width*), *H* (*height*) and *S* (*spacing*) that we can obtain W/H and S/H which inherently fixed *whse* and *whso* values and hence even and odd mode impedances were calculated for each 'bat'. The algorithm continued to find suitable values of width(*W*), height (*H*) and spacing (*S*). A limitation for $Z_{oo}$ and $Z_{oe}$ impedance values of being greater than 20Ω and less than 75Ω were also set in order to ensure realistic design values.

Some of the results were observed with the number of iterations that required to reach the optimum solution. These are shown in Table 1. It was noted that the algorithm found optimum solutions extremely fast. Most of the time, even in less than 20 iterations the approximate answer was within the 0.01 error margin to the real answer, and with a little bit of increase in iterations we observed the result to be obtained in 0.000001 error margin. Sample graphs showing the iteration convergence are shown in Figure 2 and Figure 3. Therefore when the performance metrics of speed and accuracy are considered, as it can be observed in the graphs, both of them are satisfied in a good manner.

Table 1. Results obtained using Bat Algorithm.

| Trials | W | S | H | whse | whso | $Z_{oe}$ | $Z_{oo}$ | Coupling | iteration |
|---|---|---|---|---|---|---|---|---|---|
| Trial 1 | 7.9 | 1.7 | 4.3 | 3.9 | 6.7 | 64.8 | 43.2 | 0.2 | 7 |
| Trial 2 | 11.2 | 3.6 | 7.6 | 3.3 | 5.7 | 73.2 | 48.7 | 0.2 | 99 |
| Trial 3 | 13.2 | 4.5 | 9.3 | 3.2 | 5.6 | 74.8 | 49.7 | 0.2 | 6 |
| Trial 4 | 4.1 | 0.7 | 2.07 | 4.2 | 7.2 | 61.2 | 40.8 | 0.1999 | 96 |
| Trial 5 | 17.2 | 1.2 | 5.6 | 6.3 | 10.3 | 45. | 30.2 | 0.2 | 15 |
| Trial 6 | 9.04 | 1.4 | 4.2 | 4.5 | 7.6 | 58.5 | 39.0 | 0.1999 | 102 |
| Trial 7 | 6.8 | 2.2 | 4.7 | 3.2 | 5.6 | 74.2 | 49.4 | 0.2 | 46 |
| Trial 8 | 6.3 | 1.4 | 3.5 | 3.9 | 6.6 | 65.2 | 43.5 | 0.1999 | 106 |
| Trial 9 | 6.5 | 1.9 | 4.2 | 3.4 | 6.0 | 71.0 | 47.3 | 0.2 | 36 |
| Trial 10 | 6.6 | 0.9 | 3.0 | 4.7 | 7.9 | 56.8 | 37.9 | 0.1999 | 15 |

The graphs indicate convergence to optimum answers very fast. In about 10 iterations desired coupling is obtained and in 20 iterations optimum design parameters are obtained.

International Journal of Artificial Intelligence & Applications (IJAIA), Vol. 5, No. 1, January 2014

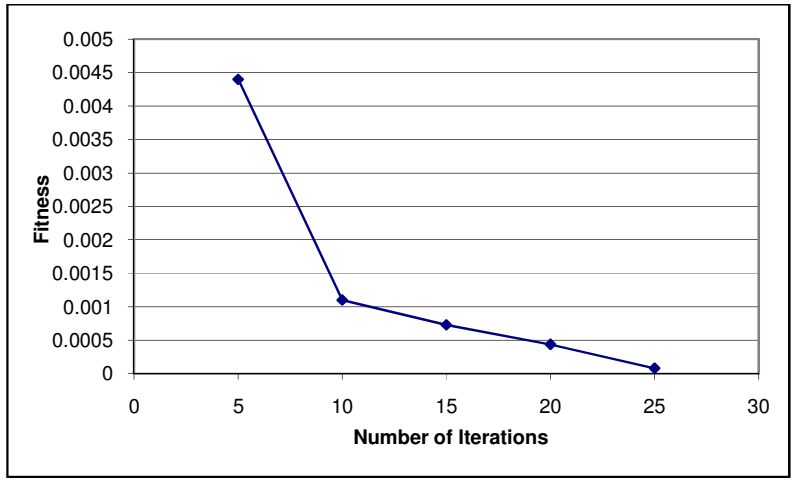

Figure 2. Convergence Graph on an independent run I

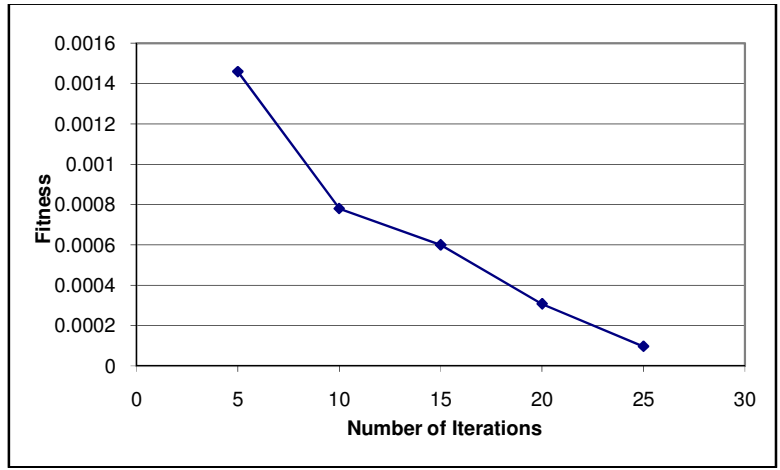

Figure 3. Convergence Graph on an independent run II

The reason for such good results is highly related with the fact the algorithm is very powerful because the algorithm uses the best of both worlds for particle swarm optimization and harmony search. Also another determining factor is the fact that there are many solutions to the problem, and we can see these different results in Table 1. When we consider the speed of the algorithm it seems that it is very powerful.

The results that were obtained should be tested and verified using microwave design simulators. This was done to make sure that the obtained values from algorithm were in fact the real design values that give the desired coupling. The verification of these data was done using a microwave simulator PUFF [14]and Sonnet microwave simulator [15], and example graph is shown in Figure 4. The figure shows the graph for the Trial 8 in Table 1. The design frequency for this problem was set to 5 GHz which was the frequency for achieving desired coupling. A smooth curve peaking at a value of 0.2 at 5 GHz was observed. The results obtained by the algorithm used in simulator indicated that they can actually be the proper design values for microwave couplers.

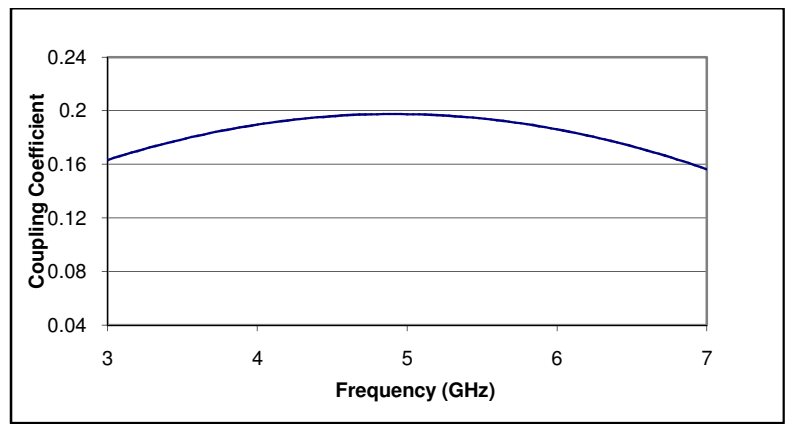

Figure 4. Coupling Coefficient and Frequency

## 5. CONCLUSIONS

In this work, the newly created bat algorithm was tested to see if it can be a possible candidate algorithm for microwave simulators in terms of speed of convergence. The algorithm was written as a program and with many different runs it produced different results. However, the results were all optimum results and the results were all found in less than 20 iterations. The example problem was microstrip coupler design equations, which can be considered as complicated functions that cannot be solved easily. The algorithm produced the dimensions for design.The obtained dimensions further were tested with microwave simulators. Simulations showed that the algorithm worked very well and the estimated dimensions by the algorithm were the proper wanted design values. The algorithm, although it seems little bit more elaborate than most of the other evolutionary algorithms, can easily be applied to the problems and good optimal design valuescan be obtained in a fast manner while designing microwave circuits.

**Authors**

**EzgiDenizUlker**

received the B.Sc., M.Sc., and PhD degree in computer engineering from Girne American University, Girne, Cyprus, in 2008, 2010 and 2013 respectively. She worked as a teaching assistant in Girne American University between 2008-2010. She worked as a lecturer in Girne American University from 2010 till 2013. She has been working as an Assistant Professor in Girne American University since December 2013. Her research interests include optimization techniques and studying different algorithms and their applications to solve different optimization and design problems. She authored and co-authored technical publications in the areas of optimization and metaheuristic algorithms. She is a member of IEEE and IEEE Computer Society.

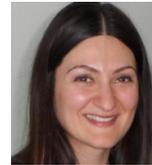

**SadikUlker**

received the B.Sc., M.E. and Ph.D. degree in electrical engineering from University of Virginia, Charlottesville, VA, USA, in 1996, 1999, and 2002 respectively. He worked in the UVA Microwaves and Semiconductor Devices Laboratory between 1996-2001 as a research assistant. Later he worked in UVA Microwaves Lab as a Research Associate for one year. He joined Girne American University, Girne, Cyprus in September 2002. He was a member of electrical and electronics engineering department as an Assistant Professor until 2009. In 2009 he was promoted to Associate Professor status and in 2013 he was promoted to Professor status in the Electrical and Electronics Engineering Department. He has also been the vice-rector of the Girne American University responsible from academic affairs since April 2009. His research interests include Microwave Active Circuits, Microwave Measurement Techniques, Numerical Methods, and Metaheuristic Algorithms and their applications to design problems. He has authored and co-authored technical publications in the areas of submillimeter wavelength measurements and particle swarm optimization technique. He is a member of IEEE and Eta Kappa Nu. He is also the recipient of Louis T. Rader Chairpersons award in 2001.

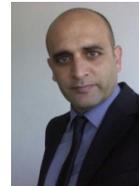